\documentclass{article}

     \usepackage[preprint]{neurips_2025}

\usepackage[utf8]{inputenc} 
\usepackage[T1]{fontenc}    
\usepackage{hyperref}       
\usepackage{url}            
\usepackage{booktabs}       
\usepackage{amsfonts}       
\usepackage{nicefrac}       
\usepackage{microtype}      
\usepackage{xcolor}         
\usepackage{graphicx}
\usepackage{tabularx,booktabs,ragged2e}
\usepackage{tcolorbox}
\usepackage[normalem]{ulem} 
\usepackage{subcaption}
\usepackage[export]{adjustbox}
\usepackage{wrapfig}
\usepackage{pifont}
\usepackage{amssymb}
\usepackage{stackengine}
\usepackage{amsmath}

\newlength{\symbolwidth}
\newcommand{\correct}{\textcolor{blue}{\ding{51}}}
\newcommand{\incorrect}{\textcolor{red}{\ding{55}}}
\newcommand{\partialcorrect}{
  \stackengine{0.68pt}{\textcolor{violet}{\ding{51}}}
  {\textcolor{violet}{\scalebox{0.9}{\ding{55}}}}{O}{c}{F}{F}{L}
}

\newcommand{\negative}{\hspace{6pt}\textcolor{red}{–}\hspace{6pt}}
\newcommand{\positive}{\hspace{6pt}\textcolor{blue}{+}\hspace{6pt}}
\newcommand{\bp}{\hspace{8pt}$\bullet$\hspace{6pt}}

\newcommand{\TODO}[1]{\textcolor{red}{--- #1}}
\newcommand{\YK}[1]{\textcolor{blue}{#1}}

\newcommand{\AD}[1]{\textcolor{red}{#1}}
\newcommand{\PT}[1]{\textcolor{brown}{#1}}
\newcommand{\ti}[1]{\textcolor{cyan}{TI: #1}}

\newcommand{\YC}[1]{\textcolor{magenta}{#1}}

\title{Defining Foundation Models for Computational Science: A Call for Clarity and Rigor}

\author{%
  Youngsoo Choi
  \\
  Center for Applied Scientific Computing\\
  Lawrence Livermore National Laboratory\\
  Livermore, CA 94550 \\
  \texttt{choi15@llnl.gov} \\
  \And
  Siu Wun Cheung \\
  Center for Applied Scientific Computing\\
  Lawrence Livermore National Laboratory\\
  Livermore, CA 94550 \\
  \texttt{cheung26@llnl.gov} \\
  \AND
  Youngkyu Kim \\
  Intelligence and Interaction Research Center \\
  Korea Institute of Science and Technology \\
  Seongbuk-gu, Seoul 02792 \\
  \texttt{youngkyu\_kim@berkeley.edu} \\
  \And
  Ping-Hsuan Tsai \\
  Department of Mathematics \\
  Virginia Tech \\
  Blacksburg, VA 24061 \\
  \texttt{pinghsuan@vt.edu} \\
  \AND
  Alejandro N. Diaz \\
  Quantitative Modeling and Software Engineering\\
  Sandia National Laboratories\\
  Livermore, CA 94551 \\
  \texttt{andiaz@sandia.gov} \\
  \And
  Ivan Zanardi \\
  Department of Aerospace Engineering \\
  University of Illinois Urbana-Champaign \\
  Urbana, IL 61801 \\
  \texttt{zanardi3@illinois.edu} \\
  \And
  Seung Whan Chung \\
  Center for Applied Scientific Computing\\
  Lawrence Livermore National Laboratory\\
  Livermore, CA 94550 \\
  \texttt{chung28@llnl.gov} \\
  \And
  Dylan Matthew Copeland \\
  Center for Applied Scientific Computing\\
  Lawrence Livermore National Laboratory\\
  Livermore, CA 94550 \\
  \texttt{copeland11@llnl.gov} \\
  \And
  Coleman Kendrick \\
  Center for Applied Scientific Computing\\
  Lawrence Livermore National Laboratory\\
  Livermore, CA 94550 \\
  \texttt{kendrick6@llnl.gov} \\
  \And
  William Anderson \\
  Center for Applied Scientific Computing\\
  Lawrence Livermore National Laboratory\\
  Livermore, CA 94550 \\
  \texttt{anderson316@llnl.gov} \\
  \And
  Traian Iliescu \\
  Department of Mathematics \\
  Virginia Tech \\
  Blacksburg, VA 24061 \\
  \texttt{iliescu@vt.edu} \\ 
  \And
  Matthias Heinkenschloss \\
  Dept.\ Computational Applied Mathematics\\
  and Operations Research,
  Rice University\\
  Houston, TX 77005 \\
  \texttt{heinken@rice.edu} \\ 
}

\begin{document}

\maketitle

\begin{abstract}
  The widespread success of foundation models in natural language processing and computer vision has inspired researchers to extend the concept to scientific machine learning and computational science. However, this position paper argues that as the term “foundation model” is an evolving concept, its application in computational science is increasingly used without a universally accepted definition, potentially creating confusion and diluting its precise scientific meaning. In this paper, we address this gap by proposing a formal definition of foundation models in computational science, grounded in the core values of generality, reusability, and scalability. We articulate a set of essential and desirable characteristics that such models must exhibit, drawing parallels with traditional foundational methods, like the finite element and finite volume methods. Furthermore, we introduce the Data-Driven Finite Element Method (DD-FEM), a framework that fuses the modular structure of classical FEM with the representational power of data-driven learning. We demonstrate how DD-FEM addresses many of the key challenges in realizing foundation models for computational science, including scalability, adaptability, and physics consistency. By bridging traditional numerical methods with modern AI paradigms, this work provides a rigorous foundation for evaluating and developing novel approaches toward future foundation models in computational science.
\end{abstract}
\section{Introduction}\label{sec:intro}\vspace{-6pt}
In recent years, there has been a surge of interest in applying machine learning (ML) and artificial intelligence (AI) to scientific and engineering domains, particularly through the lens of scientific machine learning (SciML). Alongside this growth, an alarming trend has emerged: an increasing number of researchers are labeling their models as foundation models for science and engineering—often without sufficient justification.

This trend is understandable. The success of large language models (LLMs), vision transformers, and multi-modal AI systems—commonly referred to as foundation models—has reshaped modern AI \cite{bommasani2021opportunities}. These models demonstrate broad generalization across tasks and domains, exhibit emergent capabilities, and are pretrained at scale using self-supervised learning. It is natural for researchers to want to associate their models with this success. However, the term `foundation model' carries significant scientific weight and responsibility, especially when used in a research context. 

Unfortunately, in the computational science and engineering (CSE) community, the use of the term has become ambiguous and inconsistent. Many papers now adopt the label ``foundation model'' without demonstrating the essential characteristics that would justify such a designation—such as the ability to generalize across diverse physical systems, adapt to new domains or boundary conditions with minimal or no retraining, or extrapolate meaningfully in time and space. This lack of rigor in terminology poses several risks:

\bp It dilutes the meaning of the term, eroding its value for researchers, reviewers, and practitioners.

\bp It introduces confusion in benchmarking and evaluation, especially when models are presented without proper context or generalization metrics.

\bp It undermines scientific rigor, particularly in fields that depend on reproducibility, formal guarantees, and physical validity.

In contrast, the traditional computational science community has long used the term \textit{foundational methods} to describe techniques like the finite element method (FEM) or finite volume method (FVM)—algorithms that are domain-agnostic, broadly applicable, and mathematically grounded. These methods are not data-driven, and thus not considered foundation models in the ML sense. However, they do offer a valuable reference point: a truly foundational tool in science and engineering should work across a wide range of problems with minimal reconfiguration. By analogy, a foundation model in computational science should demonstrate similar breadth—but in a data-driven setting.

Therefore, \textbf{we take the position that the term ``foundation model'' is being inconsistently and prematurely applied in computational science, and we argue that what is currently missing is a precise, field-appropriate definition of what constitutes a foundation model in computational science.} Without it, the research community lacks a shared framework for evaluating claims of generality and adaptability. This paper aims to fill that gap by:

\bp Proposing a rigorous definition of foundation models tailored to computational science;

\bp Outlining a set of core characteristics that foundation models must exhibit;

\bp Drawing connections between traditional foundational methods and modern AI-based foundation models;

\bp Helping authors, reviewers, and practitioners responsibly evaluate and categorize foundation models.

Our hope is that this framework will serve as a reference point for future work—encouraging clarity, consistency, and scientific integrity in a rapidly evolving field. We do not claim that the definition proposed in this paper is the only possible one. Rather, we hope it sparks further discussion and inspires alternative, well-reasoned definitions of foundation models tailored to computational science. What matters most is that these definitions emerge with a shared commitment to rigor, transparency, and scientific grounding.

Finally, to demonstrate the practical relevance of our proposed definition, we introduce a framework called the data-driven finite element method (DD-FEM), which is inspired by the traditional finite element method—one of the most well-established foundational methods in computational science. DD-FEM is designed to satisfy the key characteristics of a foundation model as outlined in this paper, and serves as a concrete example of how such models may be realized in practice.

\textbf{Alternative Views}: 
Several recent works have referred to their methods, such as neural operators and transformer-based surrogates, as foundation models in computational science. While these approaches demonstrate promising capabilities, such as solver acceleration, domain adaptation, or performance on specific partial differential equation (PDE) families and represent significant advancements in their respective areas, they often fall short of satisfying the broader criteria we articulate in this paper. For example, Deep Operator Network (DeepONet) \cite{lu2021learning} has been effectively used as a neural preconditioner for Helmholtz equations \cite{lee5054726ultra}, yet it does not function as a standalone general-purpose solver. Similarly, a pretrained Fourier Neural Operator (FNO) \cite{li2020fourier, pathak2022fourcastnet} has been shown to transfer across different PDEs with fine-tuning \cite{subramanian2023towards}, but the adaptation cost remains high and the generalization narrow. Poseidon, a vision-transformer-based model \cite{rabeh2024geometry}, and other neural network architectures \cite{batatia2024foundation, mendez2024mole, marcato2024developing, jakubik2023foundation} show promise in specific geometries or physics regimes, but lack demonstrations of reuse across tasks, boundary conditions, or domain topologies. These contributions are important steps forward, but they highlight the need for a clear, principled definition, one that distinguishes between high performing surrogates and models capable of broad, scalable, and reusable generalization. Our work aims to fill this gap. We elaborate this in more detail in Section~\ref{sec:reflection}.
\section{A Definition of Foundation Models in Computational Science}
\vspace{-6pt}
Examples of well-known foundation models include language models such as GPT (Generative Pretrained Transformer) \cite{vaswani2017attention, brown2020language, kalyan2024survey, bubeck2023sparks}, which have revolutionized natural language processing by capturing linguistic patterns and contexts to enable diverse downstream tasks such as text generation, translation, and sentiment analysis. In computer vision, certain models, particularly those based on large-scale pretraining and self-supervised learning, have emerged as foundation models. Notable instances include DINO (Self-\textbf{di}stillation with \textbf{no} labels)~\cite{caron2021emerging}, an example of self-supervised vision transformer, and SAM (Segment Anything Model) \cite{kirillov2023segment}, both of which demonstrate strong generalization across visual tasks like classification, segmentation, and retrieval with minimal task-specific fine-tuning. Multi-modal models, such as CLIP (Contrastive Language-Image Pre-training), Stable Diffusion, DALL-E, and Gemini, bridge vision and language modalities, enabling tasks like image generation from textual prompts and cross-modal retrieval, and exemplify the versatility of foundation models in handling multiple data formats and reasoning across modalities \cite{radford2021learning, ramesh2021zero, rombach2022high, team2023gemini}. Additionally, recent advancements such as DeepSeek \cite{guo2025deepseek, liu2024deepseek} have further expanded the capabilities of foundation models, achieving state-of-the-art performance in both language understanding and generation. These developments highlight the rapid evolution and growing sophistication of large language models, particularly in their ability to reason and generalize across tasks.

\noindent The recent surge in applying machine learning to computational science has prompted the widespread adoption of the term \textit{foundation model}—often borrowed from its successful use in natural language processing, computer vision, and multimodal AI. However, in the absence of a clear and rigorous definition tailored specifically to computational science, the term has become increasingly ambiguous. In this section, we propose a formal definition of foundation models in the context of computational science, drawing inspiration from the principles underlying traditional foundational methods, such as the FEM and FVM.

\subsection{Revisiting Foundational Methods in Classical Computational Science}\label{sec:foundational}
\vspace{-6pt}
In traditional numerical analysis, foundational methods refer to algorithms that serve as general-purpose solvers for a wide class of PDEs. For example, FEM and FVM are applicable to elliptic, parabolic, and hyperbolic PDEs across diverse geometries and boundary conditions. These methods are characterized by:

\bp\textbf{Universality and problem-independent structure}: Foundational methods are built on general-purpose numerical strategies that remain consistent across a wide range of scientific and engineering problems. Whether applied to elliptic, parabolic, or hyperbolic PDEs in fluid dynamics, solid mechanics, or electromagnetics, the core computational framework—such as weak form construction and domain discretization in FEM—remains unchanged. This universality enables modular implementation, wide reusability, and seamless adaptation to new problem settings by simply adjusting inputs such as material parameters, source terms, or boundary conditions.

\bp\textbf{Mathematical rigor}: Foundational methods are supported by a deep and mature theoretical framework. Concepts such as convergence, consistency, stability, and error estimates are precisely defined and thoroughly analyzed. This level of mathematical rigor not only ensures predictable and reproducible performance, but also provides confidence in the reliability of simulation results—which is critical for high-stakes applications in science and engineering. Additionally, these methods generally yield well-established guidelines for mesh refinement, time stepping, and solver selection, further reinforcing their robustness.

\bp\textbf{Implementation and software reusability}: A hallmark of foundational methods is their high degree of reusability at the implementation level. Once a method like FEM or FVM is implemented in software, it can be applied to a wide range of problems by simply altering the input specifications—such as geometry, mesh resolution, material properties, or boundary conditions. The underlying solver architecture and codebase do not need to be rewritten for each new application. This makes these methods particularly well-suited for general-purpose simulation frameworks, high-performance computing environments, and long-term scientific workflows, where modularity, adaptability, and scalability are essential.

These features have made such methods foundational to the field of computational science. Virtually every university that offers courses in computational science includes instruction in finite element or finite volume methods, due to their broad applicability, strong theoretical guarantees, and practical effectiveness. Moreover, these methods are not limited to academic settings: they are widely used in national laboratories, industry, and engineering practice to simulate, analyze, and predict complex physical, biological, and engineered systems. 

Despite their foundational role, these methods do not qualify as \emph{foundation models} from the perspective of AI and machine learning, as they are not data-driven. Their strength lies in first-principles numerical formulations rather than learned representations from data. However, we believe that the emerging notion of foundation models in computational science must reflect similar values, i.e., broad applicability, reusability, and robustness, achieved through data-driven learning. Building on this conceptual continuity, we now present a formal definition tailored to the needs of scientific computing.

\subsection{A Definition of Foundation Models}\label{sec:definition}
\vspace{-6pt}
Foundation models in machine learning are large-scale, data-driven models trained on broad datasets, with the ability to adapt to many downstream tasks with minimal modification \cite{bommasani2021opportunities}. To adapt this concept to computational science and to balance the general-purpose, reusable nature of traditional numerical methods with the data-driven adaptability of modern ML-based models, we propose the following definition of foundation models in computational science.

\begin{tcolorbox}[
  colback=blue!5!white,
  colframe=blue!75!black,
  title=\textbf{Definition: Foundation Models in Computational Science},
  fonttitle=\bfseries]
A \textbf{foundation model in computational science} is a data-driven model trained on a broad distribution of scientific application types or physical systems, which exhibits wide \textbf{generalization capability} across scientific problems, computational domains, tasks, and physical conditions—without requiring retraining from scratch or structural modification—and serves as a reusable base.
\end{tcolorbox}

\noindent We now clarify and justify the key phrases in this definition and share their implications:

\positive\textbf{Data-driven model}: A model is data-driven if it learns representations and mappings directly from scientific data, such as simulations, experimental measurements, or observational datasets, enabling it to capture complex, nonlinear scientific behaviors. While such a model may incorporate physical principles, its core learning mechanism must be driven by data. This distinguishes it from classical solvers like the FEM, which are not data-driven and thus do not qualify as foundation models in AI context.
    
\positive\textbf{Trained on a broad distribution of scientific application types or physical systems}: Exposure to a wide variety of physical phenomena during training allows the model to develop internal representations that are general and transferable. This breadth helps prevent overfitting to a specific problem class and sets the stage for adaptability to new, unseen scenarios—similar to the multi-task versatility seen in large language models.

\positive\textbf{Exhibits wide generalization capability across scientific problems, computational domains, tasks, and physical conditions}: A foundation model in computational science must retain strong performance when applied to new scientific problems (e.g., fluid mechanics, solid mechanics), varied computational domains (e.g., different geometries or meshes), diverse tasks (e.g., forward simulation, inverse inference, or control), and a wide range of physical conditions (e.g., different material properties, boundary conditions, or initial states). This breadth of generalization is what allows the model to be flexibly applied across disciplines and use cases without retraining from scratch.
    
\positive\textbf{Without requiring retraining from scratch or structural modification}: Once trained, a foundation model should be easily adaptable to new tasks with minimal fine-tuning and no changes to its architecture. This property ensures scalability and cost-efficiency and positions the model as a robust starting point for varied downstream workflows.

\noindent While not strictly necessary, a hallmark of foundation models in computational science is their potential to support a wide range of downstream scientific applications. These capabilities reflect desirable properties that highlight the model's practical utility and breadth of generalization, similar to the diverse downstream tasks that foundation models enable in other AI domains. Ideally, such models could accept and process multi-modal scientific data, such as various sensor measurements, simulation outputs from different PDEs or physics domains, and material descriptions. Examples of such applications include:

\bp\textbf{Single-physics simulations}: Adapting to variations in boundary or initial conditions, computational domains, or material properties within a single physical model.

\vspace{-2pt}
\bp\textbf{Multiphysics simulations}: Modeling coupled systems such as fluid-structure interaction, Darcy-Stokes flow, magneto-hydrodynamics, and electro-thermo-mechanical processes, integrating multimodal inputs and outputs across distinct physical domains. 

\vspace{-2pt}
\bp\textbf{Inverse problems and parameter estimation}: Inferring unknown system parameters or source terms from observed data or outputs, utilizing heterogeneous data including sensor readings, field data, and simulation data from different PDEs.

\vspace{-2pt}
\bp\textbf{Uncertainty quantification}: Estimating variability or confidence bounds in model predictions through probabilistic or ensemble-based techniques, integrating diverse uncertainty data such as perturbed PDE outputs, experimental noise, and material distributions.

\vspace{-2pt}
\bp\textbf{Real-time prediction and control}: Deploying the model in time-critical environments such as sensor-driven feedback loops or online control systems.

\vspace{-2pt}
\bp\textbf{Design optimization}: Using the model as a fast and accurate surrogate for high-fidelity simulations in iterative optimization workflows.

\begin{figure}
\centering
\includegraphics[width=0.9\linewidth]{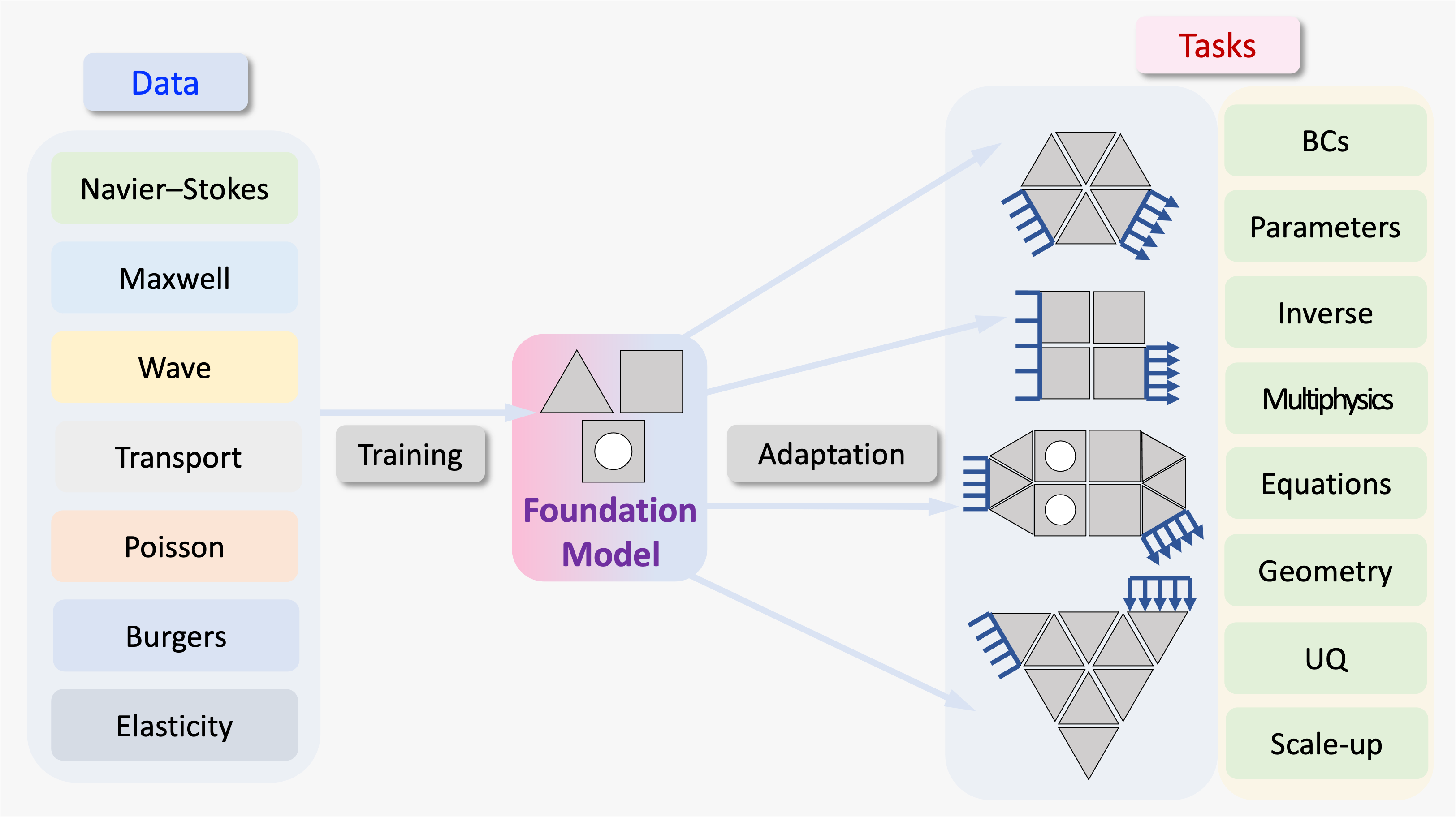}
\caption{A foundation model can consolidate knowledge from diverse problem domains into a single, unified representation, enabling it to be adapted efficiently to a wide range of downstream tasks.}
\vspace{-12pt}
\label{fig:foundationSketch}
\end{figure}

\subsection{Desirable Characteristics}\label{sec:additional}
\vspace{-6pt}
In addition to the core characteristics outlined in the definition of foundation models in computational science above, this section introduces several additional properties that, while not required, are highly desirable for models intended to serve as broadly applicable tools in scientific workflows.

\positive\textbf{Scientific Consistency}: The model should respect known physical laws or incorporate inductive biases aligned with principles such as conservation, symmetry, or invariance. Such consistency is essential for interpretability and scientific credibility, particularly in high-stakes or safety-critical domains.

\vspace{-2pt}
\positive\textbf{Robust Extrapolation in Space and Time}: While this may be inferred from generalization, it warrants explicit emphasis. A foundation model should be capable of making reliable predictions beyond its training regime, such as under new spatial domains, boundary conditions, or long-term dynamical evolution. This robustness is critical for scientific applications where test conditions often deviate significantly from those seen during training.

\vspace{-2pt}
\positive\textbf{Data Scalability}: A foundation model should not only generalize well from the data it has seen but also continue to improve as more diverse, high-quality datasets become available. Its performance should scale with data volume, allowing it to capture increasingly complex phenomena and maintain relevance as the data landscape evolves.

\vspace{-2pt}
\positive\textbf{Mathematical Rigor}: Foundation models in computational science should be supported by formal analysis where possible, such as well-posedness, stability guarantees, or convergence behavior, particularly when deployed in predictive or high-assurance settings. While full theoretical guarantees may be difficult to establish for complex models, incorporating mathematical insights can improve model robustness, guide architectural choices, and build trust in the model's scientific validity.

These characteristics draw a conceptual bridge between traditional foundational methods and modern data-driven models, helping to situate foundation models within a rigorous scientific framework.

\vspace{-6pt}
\section{Challenges in Building Foundation Models in Computational Science}\label{sec:challenges}
\vspace{-6pt}
Now that we have provided a formal definition and a set of desired characteristics for foundation models in computational science, an immediate question is whether a model meeting these criteria currently exists. It may, but the definition above definitely makes it hard to realize a foundation model in computational science for many reasons. While the success of foundation models in natural language processing and computer vision is largely enabled by standardized data formats, scalable architectures, and abundant training data, transferring these successes to computational science presents a fundamentally different set of challenges.

\negative \textbf{Massive Data Point Size}: In natural language processing (NLP) or vision, each data point (e.g., a token or image patch) is relatively small and can be processed in large batches. For example, a token in an LLM is typically embedded into a 768- or 1024-dimensional float vector, and a patch of 32\texttimes32 RGB pixels contains about 3,072 values. In contrast, a data point in computational science can easily become a vector with millions or even trillions of values, such as a 3D simulation field. These massive data sizes require gigabytes of memory per sample, severely limiting batch sizes and increasing memory pressure even on HPC-grade hardware. This leads to high variance in gradient estimates, slower convergence, and reduced statistical efficiency. Moreover, only a few thousand high-fidelity simulations may be feasible to process during training, making generalization more brittle and increasing the risk of overfitting. Finally, standard architectures like transformers or CNNs struggle to efficiently handle such high-dimensional inputs.

\vspace{-2pt}
\negative  \textbf{Expensive Fine-Tuning}: Foundation models are expected to be reusable with minimal adaptation. However, in computational science, fine-tuning is often expensive not only due to the large data sizes but also because downstream tasks may differ in PDE type, domain geometry, boundary or initial conditions, resolution, or physical regime. As a result, many parameters may need to be re-optimized, making it difficult to apply parameter-efficient transfer learning techniques like Low-Rank Adaptation of Large Language Models (LoRA) \cite{hu2022lora} or prompt tuning. This undermines the scalability and reusability that make foundation models attractive in NLP or vision.

\vspace{-2pt}
\negative  \textbf{Data Heterogeneity}: Scientific problems vary greatly in terms of physical quantities (scalars, vectors, tensors), temporal dynamics, spatial dimensionality, and governing physics. This diversity complicates the design of unified model architectures that can generalize across problem classes and physical settings.

\vspace{-2pt}
\negative  \textbf{Lack of Standardized Datasets}: Unlike NLP and vision, where large-scale public datasets (e.g., ImageNet, Common Crawl) are abundant, computational science lacks shared, high-quality datasets that span multiple physics domains. This hampers large-scale pretraining and model comparison.

\vspace{-2pt}
\negative  \textbf{Mesh and Geometry Dependence}: Scientific simulations are tightly coupled to the computational mesh and domain geometry. Variations in discretization (e.g., mesh size, topology, resolution) pose challenges for generalization and model portability across problems.

\vspace{-2pt}
\negative  \textbf{Physics-Informed Constraints}: Many scientific systems are governed by laws such as symmetry, boundary conditions, and conservation of energy. 
Incorporating these constraints into neural architectures without sacrificing generality remains a key open problem.

\vspace{-2pt}
\negative  \textbf{Trust, Interpretability, and Scientific Validity}: Scientific users often require models to be interpretable, verifiable, and consistent with physical laws. Black-box models without scientific grounding face barriers to adoption in high-stakes or safety-critical domains.

\vspace{-2pt}
\negative \textbf{Multiscale and Long-Time Behavior}: Scientific systems frequently involve multiple spatial and temporal scales. Capturing such complex dynamics in a unified foundation model—especially for long-term prediction—is extremely challenging.

These challenges underscore the need for foundational advances in architecture design, training methodology, data acquisition, and physics integration in order to realize the full potential of foundation models in computational science.

\section{A Path Toward Foundation Models: Introducing DD-FEM}
\vspace{-6pt}
In spite of the challenges outlined above, we believe that building foundation models in computational science remains an attainable goal, for example, by drawing inspiration from well-established foundational methods such as the FEM, FVM, finite difference method (FDM), and spectral methods. In this paper, we introduce the \textbf{Data-Driven Finite Element Method (DD-FEM)}, which is inspired by FEM. We believe that the new DD-FEM framework is a step toward the development of foundation models in computational science,and provide evidence through several numerical examples in Section~\ref{sec:DD-FEMenablar}.

\subsection{The DD-FEM framework: Local Learning, Global Assembly}\label{sec:dd-fem}
\vspace{-6pt}
As discussed in Section~\ref{sec:foundational}, the FEM is a foundational technique in computational science due to its universality, problem-independent structure, mathematical rigor, and implementation reusability. These characteristics make FEM a powerful and widely adopted tool across disciplines, such as structural mechanics, heat transfer, and fluid dynamics.

\begin{figure}[hbt!]
\centering
\includegraphics[width=1.0\linewidth]{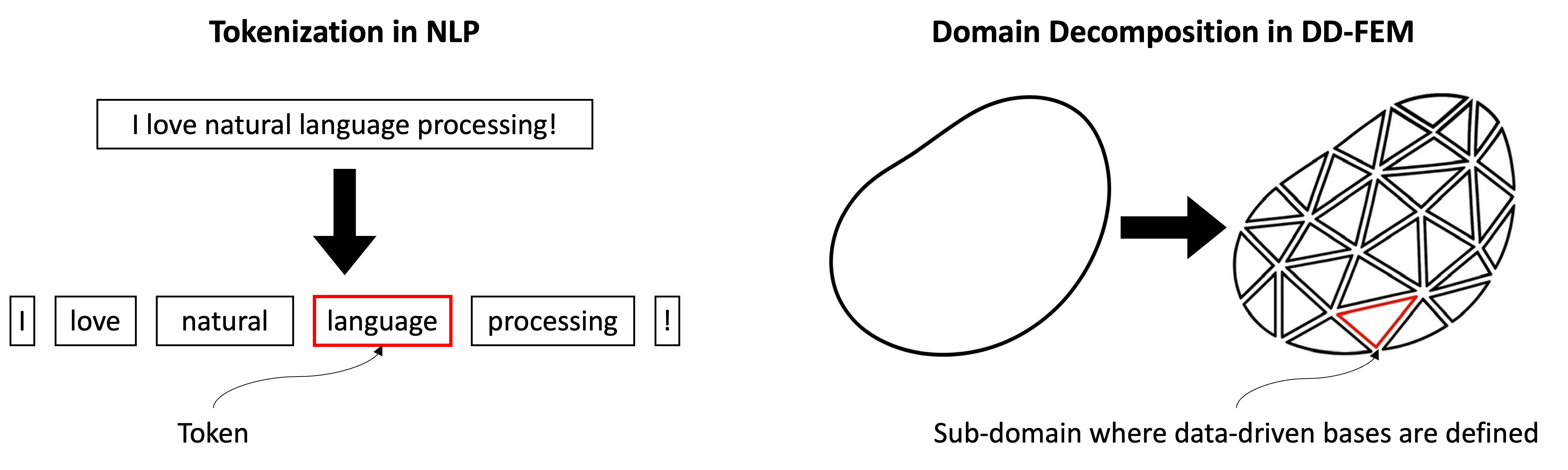}
\caption{Analogy between NLP and DD-FEM.}
\label{fig:analogyFEM-LLM}
\end{figure}

What makes FEM particularly relevant to the development of foundation models is its dual interpretation: (i) a top-down perspective in which the global domain is partitioned into smaller subdomains, and (ii) a bottom-up perspective where local basis functions are assembled to construct a coherent global solution. This brings \PT{a} natural analogy to LLMs. For example, the top-down decomposition is analogous to how language models tokenize input text (see Figure~\ref{fig:analogyFEM-LLM}), while the bottom-up assembly, driven by local interactions and physical laws, parallels the local-to-global reasoning mechanism that underlies modern foundation models, such as those based on transformers. 

\begin{figure}[hbt!]
\centering
\includegraphics[width=0.95\linewidth]{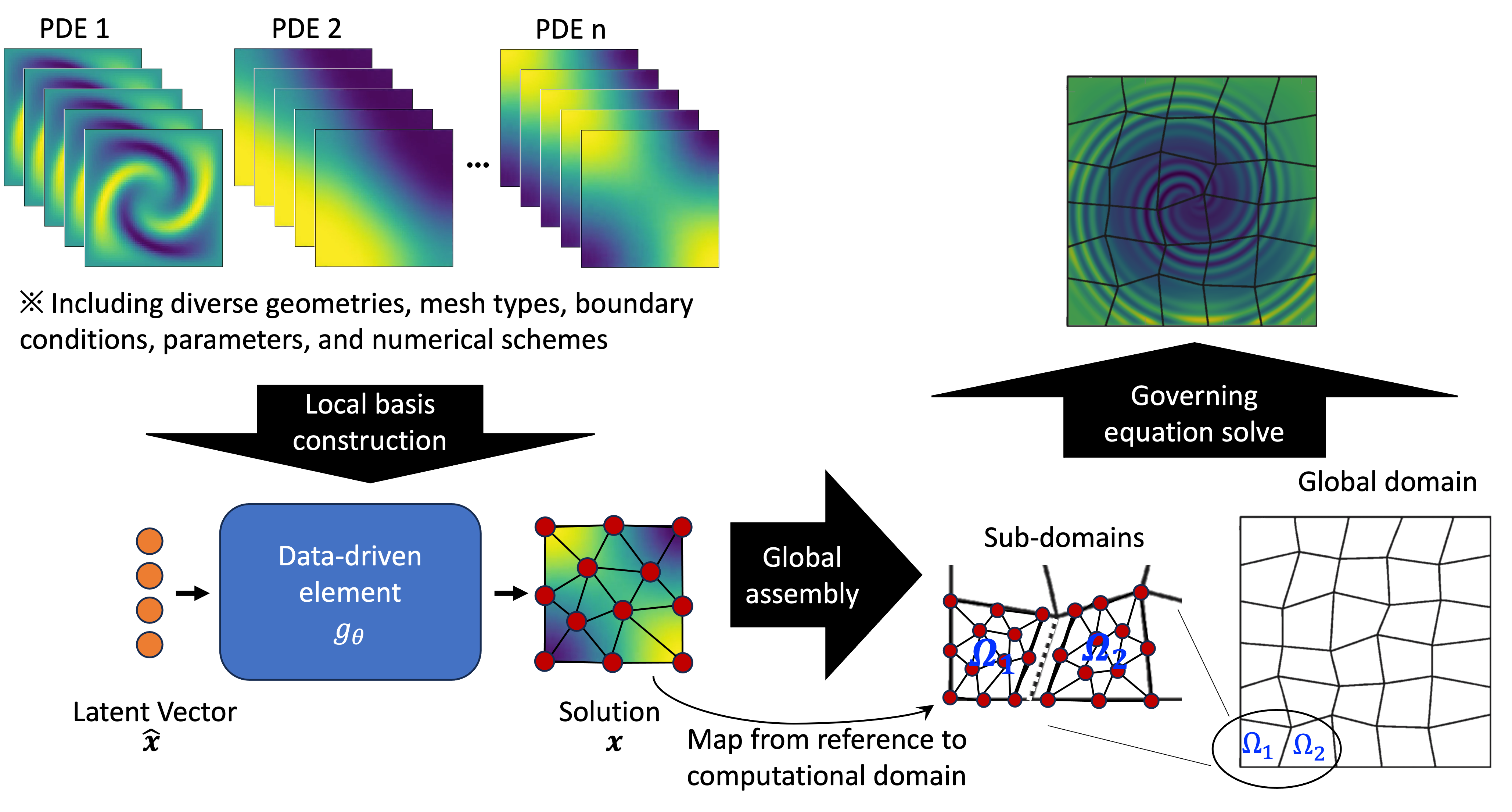}
\caption{Schematic description of the DD-FEM framework procedure: (i) Local basis construction; (ii) Global assembly; (iii) Governing equation solve.}
\label{fig:FEMprocedure}
\end{figure}

This analogy sets the stage for the DD-FEM, which retains the compositional structure of FEM while \textbf{replacing traditional polynomial bases with locally trained, data-driven bases}. Therefore, the procedure of the DD-FEM can be described as follows, which is also depicted in Figure~\ref{fig:FEMprocedure}:

\hspace{4pt}1. \textbf{Generate training data} with diverse geometries, meshes, boundary conditions, and schemes.

\vspace{-2pt}
\hspace{4pt}2. \textbf{Train local bases} on small subdomains from the generated data.

\vspace{-2pt}
\hspace{4pt}3. \textbf{Assemble global domain} using the trained data-driven elements.

\vspace{-2pt}
\hspace{4pt}4. \textbf{Solve governing equations} on the global domain to enforce physical laws.

This framework yields several important advantages. The data-driven bases offer compact and expressive representations tailored to the training data, capturing complex local behaviors more effectively than generic polynomial bases. Consequently, DD-FEM supports the use of significantly larger elements without sacrificing accuracy, allowing much larger problems to be solved within a given computational budget. Furthermore, the increased element size could contribute to greater numerical stability, permitting larger time steps and enabling longer simulation horizons compared to conventional FEM.

However, one of the most consequential advantages of DD-FEM framework in the context of foundation model development is its ability to decouple data generation from the scale of the global domain. Because its basis functions are locally defined, training data can be generated from small, computationally inexpensive subdomain simulations. This modular design enables the efficient construction of large, diverse datasets across various PDE classes without solving large-scale problems. These compact datasets can be distilled into expressive, data-driven basis functions that generalize across a broad range of physical systems. In doing so, DD-FEM directly addresses the challenge of massive data point size outlined in Section~\ref{sec:challenges} by reducing the data point size tremendously. 

The DD-FEM framework also fulfills key criteria from Section~\ref{sec:definition}, including data-driven learning and training on a broad distribution of scientific application types. Once pretrained on small-scale problems, the data-driven elements can be assembled to solve arbitrarily large systems with varying geometries, materials, boundary conditions, or initial states, without requiring retraining from scratch, provided the global domain can be constructed from the pretrained elements. This capability further aligns with the core definition of foundation models in computational science, as outlined in Section~\ref{sec:definition}. This implies robust extrapolation in space and time mentioned in Section~\ref{sec:additional}.

Furthermore, the DD-FEM framework promotes scientific consistency by numerically solving the governing equations of the underlying physics, just as FEM does, fulfilling a key criterion discussed in Section~\ref{sec:additional} and increasing trust, interpretability, and scientific validity (see Section~\ref{sec:challenges}). It also has the potential to exhibit data scalability when, for example, its data-driven bases are implemented using expressive neural network architectures, allowing the model to capture increasingly complex local phenomena as more diverse and high-quality datasets are incorporated during the training phase.

The DD-FEM is particularly well-suited to address other challenges identified in Section~\ref{sec:challenges} as well. For example, since DD-FEM operates on reusable local basis functions, adapting to new problems (e.g., with different boundary conditions or geometries) often requires updating only a small subset of local bases rather than retraining the entire model. This significantly reduces the computational burden of fine-tuning and enhances transferability.

One especially promising feature of the DD-FEM framework is its potential to enable the development of standardized datasets in scientific machine learning. By allowing data generation to be localized to small-scale subdomains, DD-FEM significantly lowers the cost and complexity of assembling high-quality datasets. This modularity encourages the creation of reusable data resources across different PDEs, geometries, and physical conditions. Collections such as The Well \cite{ohana2024well}—a diverse, large-scale benchmark suite of physics simulations—could play a pivotal role within the DD-FEM ecosystem by serving as pretraining sources for local basis functions. As more domain-specific datasets are curated, a community-driven repository of data-driven finite elements could emerge, further accelerating the development of foundation models in computational science.

A key advantage of DD-FEM in the context of foundation models is its potential to inherit mathematical rigor from classical finite element methodology, addressing one of the desirable characteristics in Section~\ref{sec:additional}. Unlike black-box machine learning models that approximate mappings without explicitly enforcing physical laws, DD-FEM numerically solves the governing equations, such as PDEs, using data-driven basis functions. This structure preserves the foundational mathematical formulation of FEM, including weak formulations, variational principles, and Galerkin orthogonality.

Because DD-FEM retains the governing equation formulation and applies data-driven approximations at the local element level, it is amenable to many of the classical theoretical tools used to analyze stability, convergence, and consistency. For example, consistency is ensured because the method still approximates solutions to the governing PDE. Stability can be analyzed in the same framework as traditional FEM, especially when numerical fluxes or stabilization techniques are used. Finally, convergence can potentially be shown under mesh refinement or enriched training datasets, building on a hybrid of FEM and surrogate modeling theory. 

Moreover, the element-wise modularity of DD-FEM makes it possible to analyze local approximation errors and propagate them to global solution bounds, opening a path toward rigorous \emph{a priori} or \emph{a posteriori} error estimates. This makes DD-FEM especially appealing for high-assurance scientific computing, where verifiability is essential.

In summary, through these properties, the DD-FEM framework exemplifies how traditional numerical structures can be hybridized with modern machine learning to build a viable pathway toward foundation models in computational science.

\section{Ways to enable DD-FEM framework}\label{sec:DD-FEMenablar}

\subsection{Compression-based}\label{sec:compression-based}
One promising approach to constructing data-driven elements within the DD-FEM framework is through compression-based modeling, following the paradigm of component reduced order models (component ROMs) \cite{huynh2013static, huynh2013static2, eftang2014port, mcbane2021component, mcbane2022stress, chung2024train, chung2024scaled, chung2024scalable, zanardi2024scalable, hoang2021domain, ebrahimi2024hyperreduced, smetana2016optimal, eftang2013port, diaz2024fast, wentland2025role}. In this approach, a compact data-driven basis is constructed by compressing high-fidelity simulation data on a discretized element domain, and the element-level governing equations are projected onto the low-dimensional manifold spanned by these bases. This procedure follows the established framework of projection-based reduced order models (ROMs) \cite{benner2015survey, choi2019space, carlberg2018conservative, hesthaven2015certified,kim2022fast, lee2020model,quarteroni2015reduced,wang2012proper,tsai2025local}, ensuring both computational efficiency and physical fidelity at the element level. These component ROMs have already demonstrated their mathematical rigor through comprehensive error analyses \cite{huynh2013static, eftang2014port, mcbane2021component}.

Component ROMs can be further categorized based on their compression type, formulation strategy, and assembly mechanism. From a compression standpoint, two primary approaches are commonly employed: linear subspace and nonlinear manifold compression. Linear subspace techniques typically use methods such as singular value decomposition (SVD) or, in the case of higher-dimensional data, tensor decompositions \cite{huynh2013static, huynh2013static2, eftang2014port, mcbane2021component, mcbane2022stress, chung2024train, chung2024scaled, chung2024scalable, hoang2021domain, ebrahimi2024hyperreduced, smetana2016optimal, eftang2013port, wentland2025role}. These approaches are computationally efficient and easier to train but may lack expressiveness in capturing highly nonlinear behavior. However, since each data-driven element only needs to capture local behavior, linear subspace techniques may be sufficient for many cases. 

In contrast, nonlinear manifold compression methods, often implemented using autoencoders, offer significantly higher representational capacity, albeit at greater training cost and complexity \cite{zanardi2024scalable, diaz2024fast}. Alternative architectures, such as generative adversarial networks (GANs), may also be employed for richer manifold learning, depending on the specific characteristics of the local solution space.  The choice between linear and nonlinear compression represents a trade-off between computational efficiency and modeling expressiveness, both of which are critical in the design of scalable and adaptable data-driven finite elements. We note/emphasize that the expressivity of the neural networks brings great potential to accomplish the generalization capability that is required by a foundation model.

The assembly of data-driven elements in the DD-FEM framework can be approached through various optimization formulations, each offering distinct advantages in terms of computational efficiency, flexibility, and enforcement of physical constraints:

\bp\textbf{Static Condensation (Null-Space Formulation)}:
Static condensation, also known as Guyan reduction, involves partitioning the degrees of freedom into internal (slave) and boundary or interface (master) sets. By expressing the internal degrees of freedom in terms of the interface or boundary ones, the system's size is reduced, leading to a condensed system that retains essential dynamic characteristics. This approach aligns with the null-space formulation, where the solution space is constrained to satisfy the governing equations implicitly, reducing the problem's dimensionality and computational cost \cite{choi2012simultaneous}. The static condensation reduced basis approaches \cite{huynh2013static, huynh2013static2, eftang2014port, mcbane2021component, mcbane2022stress, ebrahimi2024hyperreduced, smetana2016optimal, eftang2013port} follow the static condensation formulation. 

\bp\textbf{Residual Minimization with Continuity Constraints (Full-Space Formulation)}:
In this formulation, the assembly of data-driven elements is posed as a residual minimization problem, where the objective is to reduce the residual of the discretized governing equations. Continuity between elements is enforced through explicit constraints, resulting in a full-space optimization problem. Both the state (primal) and adjoint (dual) variables are treated as independent optimization variables, enabling flexible and accurate enforcement of physical laws, particularly beneficial for complex or nonlinear systems. This approach allows for systematic incorporation of constraints and generalizes well across problem settings. Representative works include \cite{zanardi2024scalable, hoang2021domain, diaz2024fast}.

\bp\textbf{Discontinuous Galerkin (DG) Methods (Soft Constraint Formulation)}: DG methods enable localized modeling by allowing discontinuities at element interfaces while enforcing continuity weakly through numerical fluxes. This strategy provides greater flexibility in handling complex geometries, non-conforming meshes, and local adaptivity. Within the optimization framework, DG can be viewed as a soft constraint formulation, where continuity across elements is not strictly enforced but penalized or controlled via flux terms. This soft enforcement enables modular assembly of data-driven elements while maintaining stability and convergence, making DG particularly attractive for scalable, component-based ROMs as DD-FEM. The soft constraint approach includes \cite{chung2024train,  chung2024scaled, chung2024scalable}.
\subsection{Collection of numerical results for compression-based approach}

\subsubsection{Demonstration of Extrapolation in Space}

\begin{figure}[ht]
    \centering
    \includegraphics[width=0.95\linewidth]{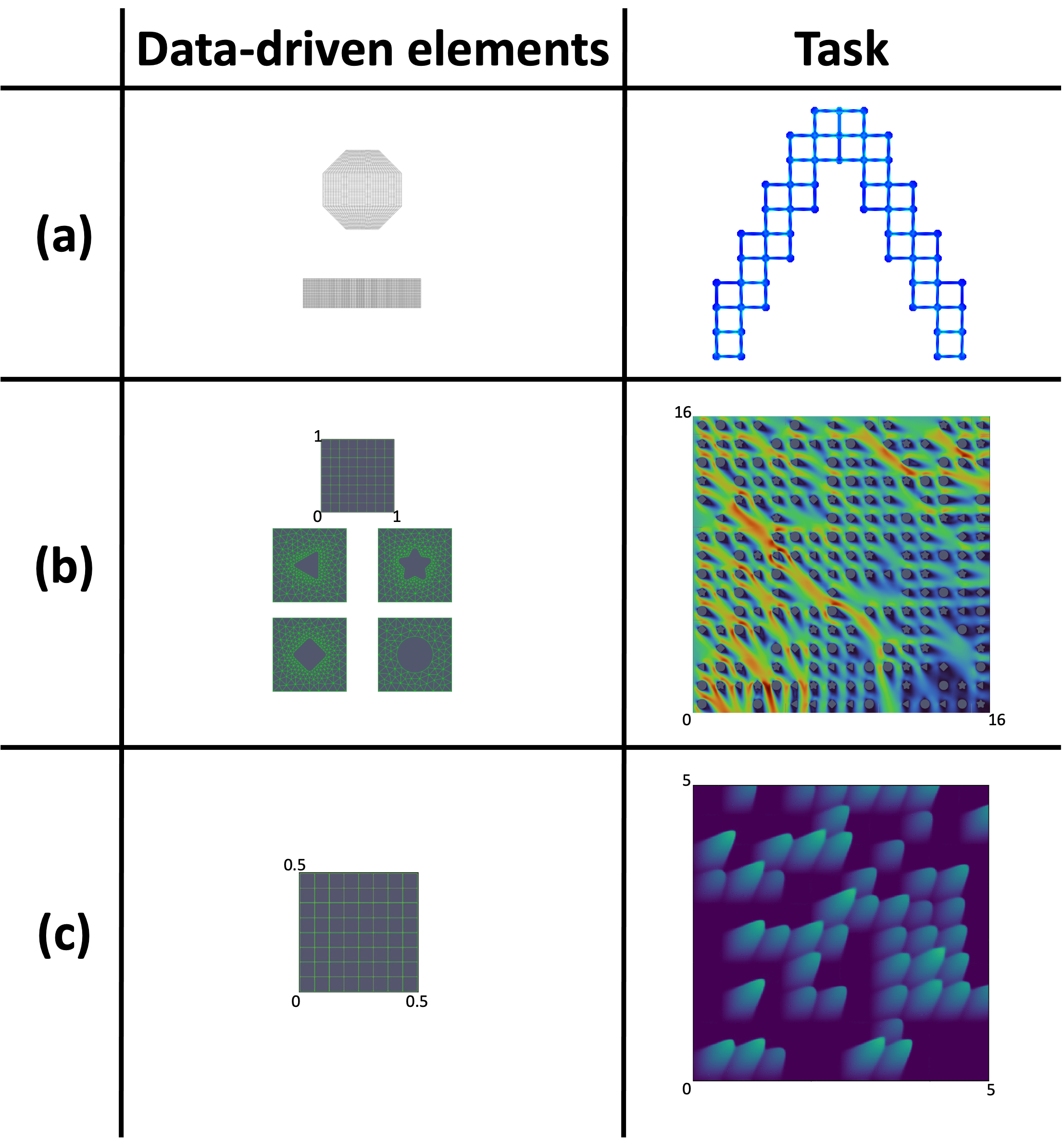}
    \caption{Demonstration of extrapolation in space. (a) lattice-type structure design optimization, (b) Navier–Stokes porous-media flow, and (c) spatially distributed disturbance Burgers flow.}
    \label{fig:extrapolationInSpace}
\end{figure}

Several component ROM approaches discussed in Section~\ref{sec:compression-based} demonstrate strong spatial extrapolation capabilities. To showcase this potential, we highlight three representative examples in Figure~\ref{fig:extrapolationInSpace}.

Figure~\ref{fig:extrapolationInSpace}(a) presents a lattice-type structural design problem composed of two data-driven elements: an octagonal joint and a strut. The underlying physics is governed by linear elasticity. Basis functions were learned using a pairwise training procedure, as described in Algorithm 2 of \cite{mcbane2021component}, and compressed via SVD. For conforming assembly, separate basis sets were constructed for interior and interface degrees of freedom, with static condensation used to enforce continuity, following a null-space formulation. This approach achieved over 1000× speedup relative to a full-order monolithic FEM solution for the global lattice structure, with less than 1\% relative error. This example illustrates DD-FEM's ability to generalize from locally trained components to unseen global geometries for a linear elasticity problem.

Figure~\ref{fig:extrapolationInSpace}(b) depicts a steady-state Navier–Stokes porous media flow problem constructed from five distinct data-driven elements with varying porous shapes. The training dataset comprises simulation snapshots generated from 2000 random 2$\times$2 subdomains, with inflow velocities sampled from a uniform distribution, as described in Eq. (44) of \cite{chung2024scaled}. As in the previous example, SVD was used for basis construction. A DG formulation was adopted, enabling soft enforcement of inter-element continuity without the need to separately construct interface and interior bases. Spatial extrapolation up to a 16$\times$16 domain was demonstrated, although the DD-FEM framework supports arbitrary scaling. The method achieved approximately 23.7× speedup with less than 4\% relative error across all scales when compared to the full-order FEM baseline. This example highlights DD-FEM’s flexibility and efficiency in generalizing to diverse and larger spatial configurations for a steady Navier–Stokes problem.

Figure~\ref{fig:extrapolationInSpace}(c) features a time-dependent 2D Burgers flow problem under spatially distributed disturbances. The domain is composed of repeated square elements, trained using data from 2000 simulations over different initial 2$\times$2 configurations with periodic boundary conditions. Unlike the previous examples, FDMs were used to generate training data. A residual minimization formulation with continuity constraints (i.e., a full-space approach) was adopted, requiring separate training of interior and interface basis functions. To effectively capture advection-dominated transient dynamics, a nonlinear manifold representation was learned using an autoencoder decoder. Spatial extrapolation to a 10$\times$10 configuration was demonstrated, yielding a 662× speedup and 1\% relative error compared to the full-order monolithic solver. This example further reinforces the scalability and generalization potential of DD-FEM across both space and time for transient, nonlinear systems. The underlying method is detailed in the recent work by Zanardi et al. \cite{zanardi2024scalable}.

\subsubsection{Demonstration of Generalization in Source Functions (Different RHS)}
\begin{figure}[ht]
    \centering
    \includegraphics[width=0.95\linewidth]{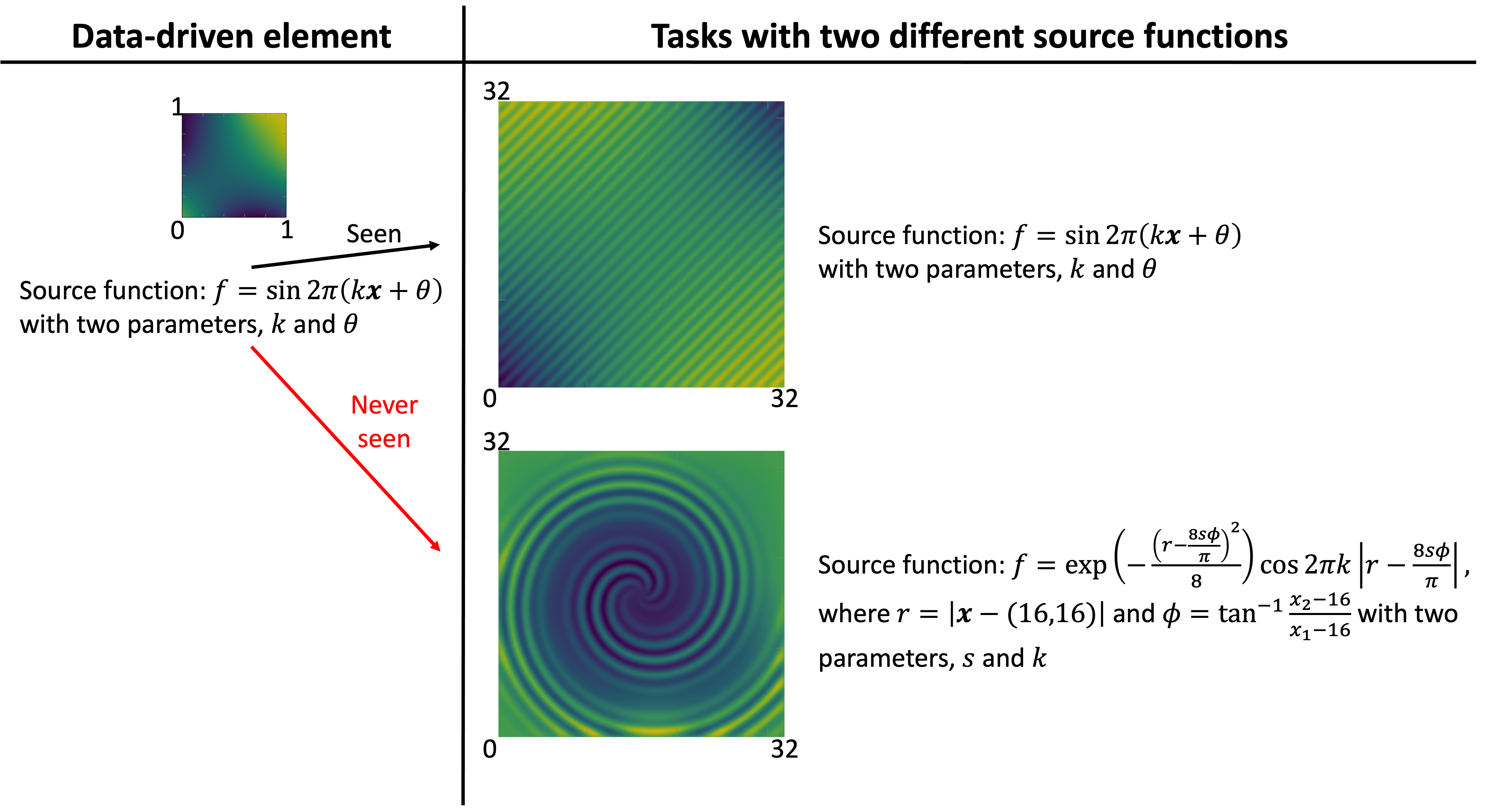}
    \caption{Demonstration of generalization to different source functions in Poisson’s equation.}
    \label{fig:generalizationSource}
\end{figure}

Figure~\ref{fig:generalizationSource} demonstrates the generalization capability of compression-based DD-FEM with respect to variation in source terms (i.e., right-hand side functions) using the 2D Poisson equation. In this example, a single type of data-driven element is trained using solutions corresponding to sinusoidal source functions of the form $f = \sin(2\pi(\mathbf{k} \cdot \mathbf{x} + \theta))$ on a unit square domain with varying frequency vectors $\mathbf{k}$ and phase shifts $\theta$ sampled uniformly from $[-0.5, 0.5]^2$ and $[0,1]$, respectively. Despite being trained solely on this class of source functions, the trained component model is then deployed in a global DD-FEM system consisting of $32 \times 32$ components, and tested on a qualitatively different source function, i.e., a spiral wave pattern never seen during training.

The spiral source is defined by a radially oscillating function centered in the domain, specified in Eq.~(33) of \cite{chung2024train}, and is significantly outside the distribution of the training data. Nonetheless, the data-driven basis needs to represent only the local region, so
the resulting DD-FEM solution remains accurate, exhibiting a relative error of approximately 0.7\% and robust qualitative agreement with the full-order solution. This outcome highlights DD-FEM’s strong inductive bias toward physical consistency and its remarkable ability to generalize from localized training to unseen global configurations, even when the right-hand side is markedly different from training inputs.

\subsubsection{Demonstration of Generalization Across PDE Types}
\begin{figure}[ht]
    \centering
    \includegraphics[width=0.9\linewidth]{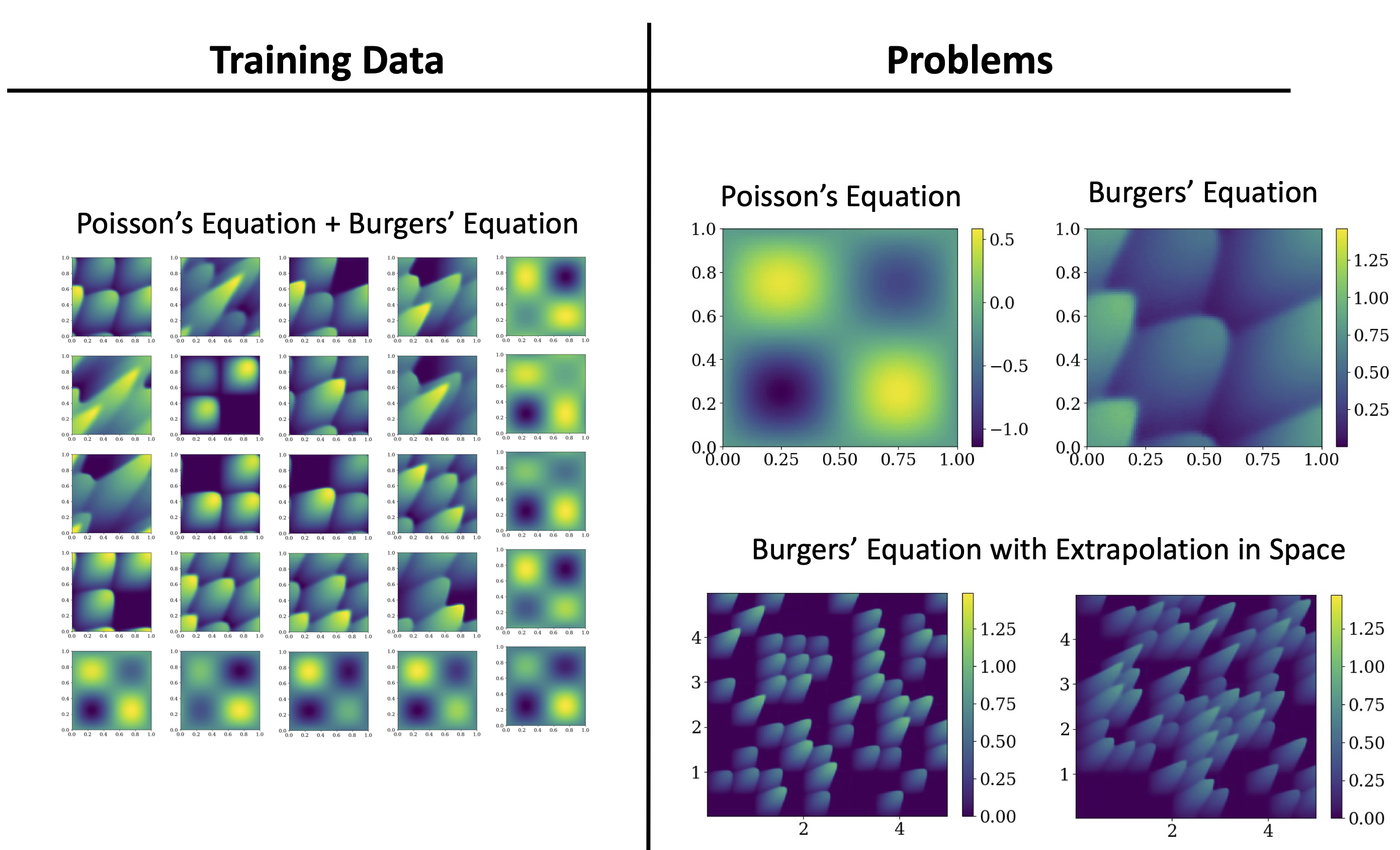}
    \caption{Demonstration of generalization to different types of PDEs.}
    \label{fig:generalizationProblem}
\end{figure}

Figure~\ref{fig:generalizationProblem} demonstrates the capability of compression-based DD-FEM to generalize across different types of PDEs. 
In this experiment, a neural decoder network was trained using a combination of 4,000 snapshots from 2D Poisson problems and 101,000 snapshots from time-dependent 2D unsteady Burgers equations with $\nu=10^{-3}$. All training data were generated from local 2$\times$2 element configurations.

Despite the differences in dynamics between elliptic and advection-dominated PDEs, the resulting compression-based DD-FEM with nonlinear manifold basis functions was able to accurately predict solutions for both types of equations under unseen testing parameters. For the Poisson problem, radial basis function (RBF) interpolation was used to generate initial guesses at the test parameters. The model achieved an average relative error of approximately 6\% for {the} Poisson {equation} and 1\% for {the} Burgers {equation}, demonstrating strong cross-problem generalization performance. Figure~\ref{fig:generalizationProblem} (top right) shows the predicted $u$-velocity field for a Poisson problem with previously unseen source terms, and 
the predicted solution for the Burgers equation with an unseen initial condition. In addition, Figure~\ref{fig:generalizationProblem}  (bottom right) shows the predicted $u$-velocity field for the Burgers equation spatially extrapolated to a $10 \times 10$ configuration, with an average relative error of 1.5\%.

While further improvements are possible, such as increasing the number of Poisson samples or tuning network hyperparameters, these results already highlight DD-FEM's promising ability to generalize across different PDE types. This example illustrates the potential for DD-FEM to support diverse scientific modeling tasks with a single data-driven basis framework, a core aspiration of foundation models in computational science.

\subsection{Other Potential Ways to Construct DD-FEM}
While compression-based DD-FEM typically employs techniques such as SVD or autoencoders to derive compact basis representations, other promising strategies may be used to construct data-driven bases once localized data is available for each element. Two noteworthy alternatives are:

\bp\textbf{Symbolic or Sparse Regression-Based Basis Construction:}  
In contrast to black-box neural networks, symbolic regression or sparse identification techniques, such as SINDy \cite{brunton2016discovering} and its weak form variant \cite{messenger2021weak}, aim to uncover compact, interpretable functional expressions from local simulation data. These methods construct basis functions by identifying a minimal set of nonlinear terms that explain the observed dynamics within each element, promoting parsimony and physical insight. The resulting symbolic models are often more transparent and analytically tractable than purely learned representations, making them especially attractive for applications that demand explainability and trust, such as high-assurance settings or safety-critical systems.

Because the resulting basis functions take on an explicit, functional form, akin to the polynomial basis functions used in classical FEM, this approach can be viewed as a data-driven analog of extended finite element methods (XFEM) \cite{moes1999finite}. In XFEM, enrichment functions are manually designed to capture discontinuities or localized features. Similarly, symbolic regression-based DD-FEM enriches the solution space with data-informed basis functions, offering a principled way to incorporate complex local behavior while preserving the mathematical structure of FEM. 

This concept also aligns with Generalized Multiscale Finite Element Methods (GMsFEM) \cite{efendiev2013generalized}, which share a common philosophy with component ROMs discussed in Section~\ref{sec:compression-based}: constructing and assembling localized basis functions. GMsFEM targets the efficient resolution of fine-scale phenomena by learning localized multiscale bases within each element, typically via local spectral problems \cite{efendiev2011multiscale}. These can be enhanced with techniques like randomized oversampling \cite{calo2016randomized} and constraint energy minimization \cite{chung2018constraint}. For global coupling, DG methods offer modularity and flexibility \cite{efendiev2013generalized-dg, cheung2020constraint}, particularly well-suited for assembling enriched elements.

This hybridization of symbolic discovery and variational approximation provides a promising direction for interpretable, generalizable, and physically grounded data-driven discretizations.

\bp\textbf{Implicit Neural Representations and Neural Operator Approaches:}  
An alternative to explicit basis construction is to use implicit neural representations or neural operators that map spatial coordinates and input parameters directly to solution values. Coordinate-based neural networks such as SIRENs \cite{sitzmann2020implicit} or FNOs can be trained per element to learn the input–output mapping in a mesh-free fashion. These representations offer several benefits: they are resolution-agnostic, easily composable, and can generalize over varying geometries and boundary conditions. Furthermore, neural operator-based formulations allow DD-FEM to move beyond projection-based frameworks, enabling non-intrusive approximations, while retaining the locality and modularity essential to DD-FEM.

\bp\textbf{Hybrid Basis Construction and Mixed Element Strategies:}  
Another flexible design choice in the DD-FEM framework is to employ mixed basis strategies, i.e., combining different types of basis functions across elements or even within a single element. For example, one may use traditional polynomial bases in regions where the solution behavior is smooth and well-understood, while leveraging data-driven bases (learned via compression, symbolic regression, or neural operators) in regions exhibiting complex, nonlinear, or data-rich dynamics. Similarly, symbolic and neural bases could be combined within an element to balance interpretability and expressiveness. Such hybridization enables tailored modeling across heterogeneous domains and allows gradual integration of data-driven techniques into legacy FEM workflows. This strategy may be especially useful during transitional deployment or in high-stakes applications where partial verification is possible only in certain regions of the computational domain.

\begin{figure}[ht]
    \centering
    \includegraphics[width=0.75\linewidth]{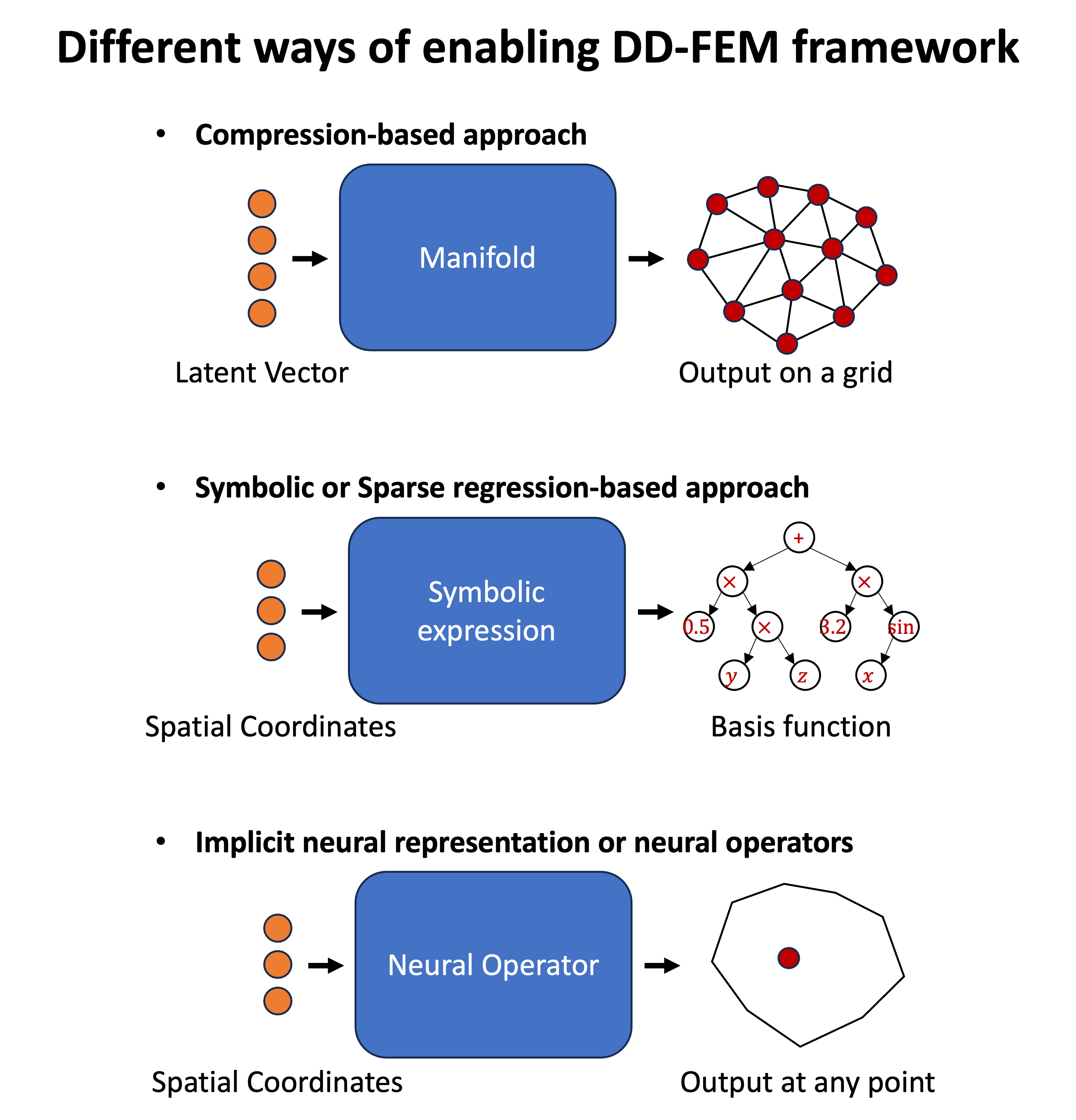}
    \caption{Three conceptual pathways to enable DD-FEM: (i) compression-based methods (e.g., SVD, autoencoders); (ii) symbolic or sparse regression; (iii) implicit neural representations or neural operators. Hybrid construction strategies can further mix and match these elements for flexible deployment.}
    \label{fig:threeDD-FEM}
\end{figure}

Each of these approaches introduces trade-offs between interpretability, generalization, training cost, and ease of integration with existing solvers. Continued exploration of these pathways will be critical for enhancing the flexibility, robustness, and adoption of DD-FEM across scientific disciplines. Figure~\ref{fig:threeDD-FEM} illustrates the three different types of DD-FEM. 

\subsubsection{DD-FEM as a Framework, Not a Specific Method}\label{sec:framework}
In summary, the DD-FEM should be understood not as a single algorithmic recipe but as a flexible framework for constructing scientific foundation models. It provides a modular blueprint for combining localized data-driven basis functions with classical numerical assembly procedures to approximate solutions to governing equations. This distinction is crucial: while most examples in this paper use projection-based ROMs to build local representations, DD-FEM is by no means limited to ROM techniques.

At its core, the DD-FEM framework consists of three conceptual building blocks:

\bp\textbf{Local basis construction:} Basis functions are trained using data from small-scale subdomains. This data may come from traditional high-fidelity solvers (e.g., FEM, FVM, FDM), experimental measurements, or synthetically generated configurations.

\bp\textbf{Global assembly:} Local basis functions are assembled into a global solution, preserving the variational structure and coupling required by the governing physics.

\bp\textbf{Physical constraint enforcement:} The assembled solution respects the underlying physical laws, ensuring scientific consistency and enabling extrapolation.

Ultimately, the power of DD-FEM lies in its openness: it defines a structure and a philosophy rather than a prescription. By decoupling the core ideas from any specific realization, DD-FEM enables continued exploration and cross-pollination between disciplines, accelerating the development of general-purpose, reusable, and physics-consistent models for computational science.

\section{Comparison among foundation model, DD-FEM, and FEM}\label{sec:comparison}
Table~\ref{ta:comparison} offers a focused comparison of foundation models in language and vision, traditional FEM, and the DD-FEM. It highlights core differences in representation, learning, and generalization, and positions DD-FEM as a bridge between classical numerical methods and modern data-driven models. By distilling complex ideas into a clear format, we hope that the table aids readers from both machine learning and computational science in understanding the novelty and potential impact of DD-FEM within the broader foundation model landscape.

Beyond this comparative insight, the hybrid nature of DD-FEM also affords extensibility in both directions:

\bp\textbf{From Traditional FEM:} Techniques such as adaptive mesh refinement, multigrid solvers, hp-adaptivity, domain decomposition, and stabilized formulations (e.g., SUPG or residual-based stabilization) can be adapted to the DD-FEM framework. These methods can help manage computational complexity, enhance numerical stability, and improve accuracy, especially in regions requiring fine resolution or exhibiting sharp gradients.

\bp\textbf{From Foundation Models:} DD-FEM can also benefit from techniques developed in the foundation model literature, including transfer learning, multi-task pretraining, continual learning, and prompt-based conditioning. These tools offer new ways to fine-tune and specialize pretrained data-driven basis functions to new PDEs, geometries, or boundary conditions with minimal retraining. Furthermore, model compression and distillation techniques may help reduce the cost of storage and inference for data-driven basis functions.

\bp\textbf{Cross-Domain Transferability:} Just as vision-language models leverage shared representations across modalities, DD-FEM could enable the reuse of pretrained elements across different PDE classes, provided appropriate representations and conditioning mechanisms are developed.

In short, the table not only clarifies conceptual distinctions among foundation models, DD-FEM, and FEM, but also highlights the extensibility and modularity that make DD-FEM a promising candidate for building foundation models in computational science.

\begin{table}
\centering
\caption{Comparison of foundation models, DD-FEM, and FEM}
\vspace{3pt}
\begin{tabularx}{\linewidth}{*{1}{>{\RaggedRight\arraybackslash}X}*{3}{>{\RaggedRight\arraybackslash}X}}
\toprule
{\bf Aspect} & {\bf Foundation models}  & {\bf DD-FEM} & {\bf Traditional FEM}  \\
\midrule
Local Units (or data points) & Data points such as Tokens (NLP), image patches (Vision), or multi-modal segments & Snapshot data in subdomains (or data-driven elements) & No data, but a finite element serves as local units, defined by spatial or temporal subdivisions of the domain \\ 
\hline
Salient feature representation $\&$ learning process & Learn functional embeddings that encode key features (e.g., semantic meaning, visual patterns, or multi-modal alignments) through pretraining & Train data-driven basis functions on small, localized domains (e.g., patches of a few elements) to capture the salient features of the solution manifold within subdomains (or data-driven elements) & Use predefined generic polynomial basis functions that are hand-engineered and problem-agnostic; no learning involved \\
\hline
Global representation & Global understanding arises from aggregating embeddings into coherent global contexts & The global domain is assembled from local elements, and the global solution is constructed by combining contributions from data-driven basis functions defined within each data-driven element & The global domain is assembled from local elements, and the global solution is constructed by combining the contributions of polynomial basis functions defined within each element \\
\hline
Principles & Principles like grammars, contextual relationships, or self-attention mechanisms guide the assembly of tokens into coherent structures (e.g., sentences, paragraphs, complete images) & Physics laws (e.g., conservation laws, equilibrium equations) govern the assembly of data-driven element contributions into a physically consistent and accurate global solution & Physics laws (e.g., conservation laws, equilibrium equations) govern the assembly of local element contributions into a physically consistent and accurate global solution \\
\hline
Generalization & Pretrained models can generalize to diverse tasks or genres with or without minimal fine-tuning (e.g., GPT can generate essays, code, and poetry); Any global representation that can be made out of pre-trained functional embeddings & DD-FEM can solve a wide range of physics and engineering problems on any global domain constructed from pretrained data-driven bases defined on a set of data-driven elements. Broad generalization is achievable through fine-tuning, enabling adaptation of these pretrained bases to new physical systems or problem settings & FEM can solve virtually any physics or engineering problem across a wide range of global domains, boundary conditions, and initial conditions  \\
\bottomrule
\end{tabularx}
\label{ta:comparison}
\end{table}

\section{Implementation outlook for DD-FEM}\label{sec:implementation}
In this section, we briefly discuss the key components of a DD-FEM implementation, as well as several key principles to follow when implementing DD-FEM.
At a high level, a DD-FEM implementation will need to define several key components: the construction of small subdomain problems (training), the formation of a global domain, and the solution on the global domain.

\bp\textbf{Construction of subdomain problems:} For a given class of computational domains, we construct a representative set of subdomains partitioning the global domain. These small subdomains may vary in their geometry, material properties, boundary conditions, source functions, etc. Their size should be limited, to ensure efficient construction of local bases with low dimension. 

\bp\textbf{Formation of a global domain:} Using the subdomains as building blocks, the global domain is assembled. The subdomains form a geometric partition of the global domain, and they are chosen to reproduce the global system with correct material properties, boundary conditions, etc. The formulation of the global system also requires coupling conditions, which is a problem-dependent choice. In the FEM context, a continuous Galerkin formulation may use interface bases to ensure continuity, whereas a discontinuous Galerkin approach may use weak enforcement. For validation of the DD-FEM solution, a full-order discretization at high resolution may also be implemented, for error computations on systems of modest size.

\bp\textbf{Solving the global system:} The resulting system on the global domain should be much smaller than that of a full-order discretization, but the system size may still be formidable for large problems. The global system is analogous to a high-order FEM system, in that the interaction of basis functions globally is limited to neighboring subdomains, resulting in a block-sparse system. The usual challenges of scalable, preconditioned linear solvers must be addressed, depending on the type of system.

\bp\textbf{Scalability:} Even though DD-FEM is computationally efficient in the construction phase due to the use of small subdomains, the assembly of a large number of subdomains can lead to significant bottlenecks during the global system solve. To ensure scalability, parallel implementation is essential. Subdomains and their corresponding system blocks can be distributed across processors using MPI parallelization, and GPU acceleration can further improve throughput for both training and inference phases. Additionally, the global DD-FEM system, while smaller than full-order models, is often high-order and block-sparse, and therefore demands the development of effective preconditioners. Preconditioners tailored to the unique structure of DD-FEM (e.g., modular local bases, heterogeneous physics, and coupling formulations) will be crucial for achieving efficient and scalable solver performance, particularly as problem sizes grow in both resolution and complexity.

\bp\textbf{Ease of use:} Using configuration files can provide a convenient way for users to define the components of the global domain, choose between FEM types (such as mesh geometries, boundary conditions, etc.), and specify various system parameters for both the training and testing phases. Tools for visualization and computing quantities of interest, globally or on local subdomains, may also be valuable.

\bp\textbf{Reproducibility:} Ensuring reproducibility is essential for the credibility and long-term utility of DD-FEM research. To achieve this, implementations should adopt version-controlled repositories with clear open-source licenses and maintain comprehensive documentation of dependencies, system requirements, and software configurations. Training pipelines should include scripts for data generation, preprocessing, model training, and evaluation, along with accessible archives of the corresponding datasets and metadata, such as subdomain geometries, boundary conditions, and random seeds. Moreover, implementations should include logging capabilities that record hyperparameters, mesh resolutions, loss histories, and solver tolerances. These measures collectively enable researchers to replicate results, verify performance claims, and extend the methodology in a transparent and scientifically rigorous manner.

These practices will ensure that experiments can be reliably replicated, results can be independently verified, and future researchers can build on prior work with confidence.

\section{Clarification on the Naming of Data-Driven Finite Element Method (DD-FEM)}\label{sec:clarification}

We refer to our proposed framework as the \textit{Data-Driven Finite Element Method (DD-FEM)} because it conceptually extends the classical finite element method by replacing generic polynomial basis functions with locally trained, data-driven bases. This naming highlights the framework’s structural similarity to FEM and its compositional, element-wise assembly.

However, as discussed in Section~\ref{sec:DD-FEMenablar}, the data used to train these bases need not originate from FEM simulations specifically. It can come from any traditional discretization method, such as finite volume (FVM) or finite difference (FDM). Moreover, the assembly procedure in DD-FEM is not restricted to a single strategy; it supports a range of formulations, including static condensation, residual minimization, and discontinuous Galerkin approaches, and can operate in both discrete and continuous settings.

Thus, while the name DD-FEM reflects its inspiration and modular architecture, the framework itself is broadly applicable. Alternative names, such as \textit{Data-Driven Finite Basis Method (DD-FBM)} or \textit{Data-Driven Discretization Method (DD-DM)}, might better capture this generality. Nonetheless, we retain the DD-FEM nomenclature in this work to emphasize its conceptual lineage with FEM, while clarifying that the core contribution, i.e., data-driven basis construction, is flexible and extensible across multiple discretization paradigms in computational science and engineering.
\section{Reflections on Models Claimed as Foundation Models}\label{sec:reflection}

As discussed in Section~\ref{sec:intro}, the term ``foundation models'' has gained rapid popularity, and it is increasingly applied to a broad range of machine learning approaches in computational science. Now that we have defined the meaning of foundation models in a thorough way, it is worthwhile to revisit recent works that self-identify or are referred to as foundation models through the lens of our definition. It is important to note that our aim here is not to diminish the novelty or scientific contribution of these works, but rather to evaluate their alignment with the rigorous criteria for a true foundation model in computational science as established in this paper.


We are not the first to explore what constitutes a foundation model in computational science. A notable early effort is \cite{subramanian2023towards}, which investigates whether FNOs pre-trained on a range of PDE systems can serve as reusable bases for scientific modeling. 
Their study translates the ``pretrain-and-finetune'' paradigm from NLP and vision into SciML, and makes a significant contribution by demonstrating transfer learning across three diverse linear elliptic PDEs -- Poisson, advection–diffusion, and Helmholtz -- within a unified training pipeline. The model is trained on a mixed dataset incorporating each PDE class and leverages structured zero-channel input padding to prompt operator-specific predictions without requiring architectural modification. Fine-tuning from this mixed model improves downstream performance in both zero-shot and few-shot settings, highlighting strong intra-domain transferability. These findings satisfy several key aspects of our definition, including training on diverse physical systems, model reuse without structural changes, and data scalability. However, the model’s generalization remains limited to surrogates within the class of linear elliptic PDEs, and it does not incorporate explicit physical constraints such as conservation laws, symmetry, or energy principles during inference.

In \cite{lee5054726ultra}, DeepONets are used within a hybrid preconditioning strategy for accelerating iterative solvers in parametrized PDEs. The authors refer to these networks as foundation models due to their independence from specific scatterer geometries and their extrapolation across shape variations. While this demonstrates promising reusability within a narrow task class, the model is not applied beyond preconditioning, lacks scientific consistency and does not generalize across different scientific problems or modeling tasks, limiting its scope relative to the criteria of a foundation model in computational science.

Many existing foundation models are domain-specific. One such example is \cite{batatia2024foundation}, where the authors develop MACE-MP-0, a graph neural network trained on over 150,000 inorganic crystal structures, and frame it explicitly as a foundation model for atomistic materials chemistry. The model is applied across a wide spectrum of applications—ranging from molecular dynamics of solids, liquids, and gases to catalysis and metal–organic frameworks—without requiring retraining. While MACE-MP-0 demonstrates exceptional generalization within atomistic modeling and satisfies many core criteria for foundation models, it does not incorporate explicit scientific laws such as conservation laws or symmetries, and its scope remains limited to the atomistic simulation regime.

An additional reason some works are identified as foundation models is the use of large language models or vision transformers. For example, MolE, a molecular foundation model proposed in \cite{mendez2024mole}, adapts the DeBERTa architecture \cite{he2020deberta} and is pretrained on large-scale molecular and biological datasets. It is fine-tuned for downstream drug discovery tasks, capturing the essence of a foundation model through task diversity and transfer within a single modality. However, the model remains focused on molecular graphs and does not demonstrate generalization across different scientific domains such as materials science or biophysics. Additionally, the architecture lacks built-in physical constraints or inductive priors, which limits its scientific consistency under the definition proposed. Text-based molecular foundation models are increasingly supported by infrastructure like Smirk \cite{wadell2024smirk}, a tokenizer that ensures chemically complete and scalable molecular representations. 

In \cite{rabeh2024geometry}, the authors refer to vision transformer–based models as foundation models for fluid dynamics due to their improved generalization in data-scarce regimes and across shape variations. The study highlights the scalability and adaptability of such models within a family of flow problems, suggesting a degree of reusability. However, the model is benchmarked solely within the domain of steady-state incompressible flow, and generalization to unrelated physical systems or computational tasks is not demonstrated. Moreover, the approach does not inherently incorporate or guarantee scientific consistency, limiting its alignment with the broader definition of foundation models in computational science.

In \cite{marcato2024developing}, the authors describe their work as the first foundation model for predicting material failure, trained on data from rule-based, phase-field, and finite-discrete element simulations across a range of materials. The model handles structured and unstructured grids, multiple fracture prediction tasks, and context embeddings from a large language model, enabling a degree of zero-shot generalization. It also demonstrates favorable scaling behavior and reusability across input formats without structural modification. However, the model lacks explicit enforcement of physical principles such as conservation laws or symmetry, and its generalization remains confined to the domain of fracture mechanics. While promising, it only partially satisfies the criteria for a foundation model in computational science as defined in this work.

Another example is \cite{jakubik2023foundation}, which introduces a transformer-based geospatial foundation model pretrained on over 1 TB of multispectral satellite imagery and designed for multi-task Earth observation. The model is fine-tuned for flood mapping, fire scar segmentation, cloud imputation, and crop type identification, demonstrating broad reusability, strong scaling behavior, and data efficiency within the remote sensing domain. The model meets many of the key criteria for foundation models, including broad training data, diverse downstream utility, and architectural reusability. However, it does not incorporate scientific constraints, and its generalization remains within the realm of satellite imagery and geospatial analysis rather than spanning across scientific modalities. Nonetheless, it represents one of the most complete realizations of a domain-specific foundation model in computational Earth science to date.

As summarized in Table~\ref{tab:foundation-model-comparison}, many existing models labeled as foundation models in computational science are often trained primarily on empirical data without explicitly incorporating physics-based constraints. This absence often results in predictions that violate core physical principles—such as conservation laws, symmetry conditions, or energy consistency—and undermines their ability to generalize in space, time, or across scientific domains. Furthermore, approaches like DeepONet and FNOs, currently possess relatively limited theoretical frameworks for error reduction and stability guarantees in practical applications. Consequently, these models may lack the robustness, scientific consistency, and domain transferability required by the definition of foundation models proposed in this paper.

To overcome these fundamental limitations, we advocate for hybrid approaches that seemlessly integrate data-driven learning with established physics-based modeling frameworks. One such example is the Data-Driven Finite Element Method (DD-FEM) introduced in Section~\ref{sec:dd-fem}. DD-FEM constructs reusable, physically consistent bases from small-scale problems while preserving the governing equations at the global scale. This combination of local learning and global structure offers a principled way to embed physical constraints and mathematical models into foundation model architectures. When used alongside purely data-driven models, DD-FEM can serve as a complementary scaffold, closing the gap between generalization and scientific rigor. As shown in Table~\ref{tab:foundation-model-comparison}, DD-FEM satisfies key foundation model criteria, including space-time extrapolation, reusability, scientific consistency, and mathematical rigor.


\begin{table}[ht]
\centering
\renewcommand{\arraystretch}{1.2}
\begin{tabular}{lcccccccc}
\toprule
\textbf{Criteria} & \textbf{\cite{subramanian2023towards}} & \textbf{\cite{lee5054726ultra}} & \textbf{\cite{batatia2024foundation}} & \textbf{\cite{mendez2024mole}} & \textbf{\cite{rabeh2024geometry}} & \textbf{\cite{marcato2024developing}} & \textbf{\cite{jakubik2023foundation}} & DD-FEM \\
\midrule
\textbf{Training Distribution} & \partialcorrect & \partialcorrect & \correct & \correct & \correct & \correct & \correct & \correct \\
\textbf{Cross-domain Generalization} & \correct & \incorrect & \partialcorrect & \incorrect & \incorrect & \incorrect & \partialcorrect & \correct \\
\textbf{Reusability} & \partialcorrect & \incorrect & \correct & \correct & \partialcorrect & \correct & \correct & \correct \\
\textbf{Space-time Extrapolation} & \partialcorrect & \incorrect & \correct & \partialcorrect & \partialcorrect & \correct & \correct & \correct \\
\textbf{Data Scalability} & \correct & \incorrect & \correct & \correct & \correct & \correct & \correct & \correct \\
\textbf{Scientific Consistency} & \incorrect & \incorrect & \incorrect & \incorrect & \incorrect & \incorrect & \incorrect & \correct \\
\textbf{Mathematical Rigor} & \incorrect & \incorrect & \incorrect & \incorrect & \incorrect & \incorrect & \incorrect & \correct \\
\bottomrule
\end{tabular}
\vspace{6pt}
\caption{Comparison of existing works on foundation models in computational science and DD-FEM against the criteria in Section~\ref{sec:definition} and Section~\ref{sec:additional}. Symbol definition: Satisfied: \correct; Partially Satisfied: \protect\partialcorrect; Not Satisfied: \incorrect.}
\label{tab:foundation-model-comparison}
\end{table}

\section{Discussion and Outlook}
\vspace{-6pt}
\textbf{Clarity and Rigor in Terminology.}
As foundation models gain traction in scientific machine learning, their application in computational science risks becoming ambiguous without a clear and rigorous definition. In this paper, we have proposed a scientifically grounded definition rooted in generalization, reusability, and data-driven adaptability. Our goal is to provide a principled reference point for researchers, reviewers, and practitioners, helping to anchor future developments in clarity and scientific rigor.

\textbf{DD-FEM as a Path Forward.}
We introduced the Data-Driven Finite Element Method (DD-FEM) as a promising architectural framework for building foundation models in computational science. By replacing traditional polynomial basis functions with data-driven alternatives, DD-FEM combines the modularity, local-to-global structure, and mathematical rigor of FEM with the flexibility of modern representation learning. Its support for localized training and compositional assembly makes it adaptable to diverse problem configurations and compatible with multiple discretization paradigms.

\textbf{Bridging Disciplines and Enabling Standardization.}
DD-FEM offers a unique opportunity to bridge the gap between classical numerical methods and data-driven modeling. Its reliance on local training problems enables inexpensive data generation, laying the foundation for standardized, community-driven datasets and benchmarks. Existing modular repositories, such as the Well dataset, can be extended within the DD-FEM framework to promote reproducibility, sharing, and reusability across domains.

\textbf{Open Questions for Community Discussion.}
This position paper is intended not as a final declaration, but as a catalyst for broader community dialogue around the definition, scope, and responsible development of foundation models in computational science. While we propose a rigorous definition and offer a concrete framework through DD-FEM, we recognize that the landscape is still evolving and that alternative perspectives must be explored. Open questions remain regarding the boundaries of generalization, the role of physical priors, and the extent to which empirical performance versus theoretical guarantees should govern foundation model status. We invite the community to reflect on the following questions as we collectively shape the standards and benchmarks for this emerging field.

\bp Although DD-FEM has strong potential to enable foundation models in computational science, what other frameworks could also satisfy the definition and characteristics proposed in this paper?

\vspace{-2pt}
\bp What other definitions of foundation models could apply to computational science?

\vspace{-2pt}
\bp Should scientific consistency, e.g., respect for conservation laws, be a strict requirement, or can foundation models be effective even when violating known physics?

\vspace{-2pt}
\bp How should we benchmark, certify, and track progress toward general-purpose, reusable models for computational science?

\vspace{-2pt}
\bp What types of datasets and evaluation tasks best reflect the versatility and reliability expected of foundation models in this field?

\vspace{-2pt}
\bp Can a domain-specific model (e.g., only fluid or solid mechanics) still qualify as a foundation model if it exhibits sufficient generalization and reusability within that domain?

\textbf{Looking Ahead.}
By defining foundation models for computational science and introducing DD-FEM as a concrete, extensible framework, we aim to spark a principled discussion on building general-purpose, reusable, and scientifically consistent models. These contributions are not intended as final answers, but as starting points for further dialogue and innovation. As numerical methods and machine learning converge, we invite the community to engage, question, and expand on this vision.

\section{Open Research Directions for DD-FEM}
While DD-FEM represents a significant step toward realizing foundation models in computational science, several open challenges remain:

\bp\textbf{Generalization across subdomain geometries and physics:}
How can we develop representation learning methods that generalize across arbitrary mesh structures, subdomain shapes, boundary conditions, and PDE types? The isogeometric mapping between reference and computational domains, commonly employed in traditional FEM (see Figure~\ref{fig:FEMprocedure}), can also be leveraged in DD-FEM, provided a topological isomorphism exists between the domains. However, for more complex geometries, rapid adaptation to new subdomains remains a key challenge. We believe this is achievable due to DD-FEM’s inherently localized training paradigm, which supports efficient retraining on newly encountered subdomains.

\bp\textbf{Inductive biases and neural design:} What architectural choices or physics-informed priors best capture local physical behavior within data-driven bases? Identifying the right inductive biases, such as symmetry preservation, conservation laws, or locality constraints, is critical for ensuring that learned basis functions respect the underlying physics and generalize across problem settings. Neural architectures tailored to PDEs, such as FNOs, or graph neural networks, may offer promising avenues for encoding these biases. Additionally, the interplay between model capacity and physical fidelity must be carefully balanced: overly expressive models may overfit to training data, while overly constrained ones may fail to capture key dynamics. Future work should explore how domain knowledge can guide architecture design, loss functions, and training procedures to yield efficient, generalizable, and physically meaningful representations within the DD-FEM framework.

\bp\textbf{Mathematical guarantees:} Although DD-FEM retains the governing structure of FEM, the introduction of data-driven bases raises questions about new sources of approximation error. Can we extend classical FEM and domain decomposition theory to rigorously characterize convergence, stability, and error propagation in DD-FEM? Formalizing such guarantees is essential to achieving the level of trust long associated with traditional foundational methods.

\bp\textbf{Flexible assembly strategies:}
As discussed in Section~\ref{sec:DD-FEMenablar}, DD-FEM supports a variety of coupling strategies, including static condensation, residual minimization, and discontinuous Galerkin formulations. Understanding the trade-offs among these approaches, in terms of stability, accuracy, computational efficiency, and ease of implementation, remains an important area for future research. Moreover, DD-FEM's modular structure naturally enables hybrid assembly strategies that combine data-driven and classical discretizations. For instance, DD-FEM elements can be seamlessly coupled with conventional FEM elements or integrated into existing simulation pipelines and legacy codes. This opens the door to incremental adoption in established workflows and supports heterogeneous modeling environments where data-driven components can be selectively applied to regions of high complexity or uncertainty.

\bp\textbf{Implementation and software reusability:} The integration of data-driven elements poses new challenges in model deployment and system design, including storing learned bases, ensuring compatibility across discretizations, and interfacing with solvers. Developing reusable, scalable infrastructure, complete with standardized APIs and shared model libraries, will be essential for widespread adoption.

\textbf{Broader Impacts.} The DD-FEM framework has the potential to democratize and accelerate scientific modeling in high-impact domains such as climate science, 
materials discovery, and energy systems. By enabling standardized, reusable, and interpretable data-driven components grounded in physical laws, it supports a trustworthy and responsible vision of AI for science, one that values reproducibility, collaboration, and scientific integrity.

\begin{ack}
This work was supported by the U.S. Department of Energy (DOE), Office of Science, Office of Advanced Scientific Computing Research (ASCR), through the CHaRMNET Mathematical Multifaceted Integrated Capability Center (MMICC) under Award Number DE-SC0023164 to YC. YC, Seung Whan C, and DMC also received support from ASCR as part of the Advancements in Artificial Intelligence for Science program.
The Lawrence Livermore National Laboratory (LLNL) Laboratory Directed Research and Development (LDRD) Program provided additional support under Project No. 24-SI-004 to YC and Seung Whan C, and Project No. 24-ERD-012 to YC and Siu Wun C.
TI was supported by the National Science Foundation under Grant DMS-2012253. AND received funding through the S. Scott Collis Fellowship at Sandia National Laboratories.
Livermore National Laboratory is operated by Lawrence Livermore National Security, LLC, for the U.S. Department of Energy, National Nuclear Security Administration under Contract DE-AC52-07NA27344. LLNL document release number: LLNL-CONF-2006534.
\end{ack}

\bibliographystyle{unsrt} 
\bibliography{references}

\end{document}